\documentclass[conference, twocolumn]{IEEEtran}
\usepackage{microtype}
\usepackage[hyphens]{url}

\usepackage{braket}
\usepackage{listings}
\usepackage{graphicx}
\usepackage[table,xcdraw]{xcolor}
\usepackage{amsmath}
\usepackage{pifont}
\usepackage{amssymb}
\definecolor{mydarkblue}{rgb}{0,0.08,0.65}
\usepackage[colorlinks=true,
    linkcolor=mydarkblue,
    citecolor=mydarkblue,
    filecolor=mydarkblue,
    urlcolor=mydarkblue]{hyperref} 
\usepackage{cleveref}
\usepackage{soul}
\usepackage[shortlabels]{enumitem}
\usepackage{tikz}
\usepackage{balance}
\usepackage{multirow}
\usepackage{comment}
\usepackage{wrapfig,lipsum,booktabs}

\usepackage[symbol]{footmisc}
\usepackage{tablefootnote}
\usepackage{tabularx}
\usepackage{placeins}
\usepackage{algorithm}
\usepackage{algpseudocode}
\usepackage{makecell}
\usepackage{cuted}
\usepackage{float}
\usepackage{caption}

\usepackage{multicol}

\usepackage{dblfloatfix}
\usepackage{natbib}

\renewcommand{\arraystretch}{1.2}

\definecolor{codegreen}{rgb}{0,0.6,0}
\definecolor{codegray}{rgb}{0.5,0.5,0.5}
\definecolor{codepurple}{rgb}{0.58,0,0.82}
\definecolor{backcolour}{rgb}{0.95,0.95,0.92}

\makeatletter
\def\blfootnote{\xdef\@thefnmark{}\@footnotetext}
\makeatother

\lstdefinestyle{mystyle}{
  backgroundcolor=\color{backcolour},   commentstyle=\color{codegreen},
  keywordstyle=\color{magenta},
  numberstyle=\tiny\color{codegray},
  stringstyle=\color{codepurple},
  basicstyle=\ttfamily\footnotesize,
  breakatwhitespace=false,         
  breaklines=true,                 
  captionpos=b,                    
  keepspaces=true,                 
  numbers=left,                    
  numbersep=5pt,                  
  showspaces=false,                
  showstringspaces=false,
  showtabs=false,                  
  tabsize=2,
}

\AtBeginDocument{%
  \providecommand\BibTeX{{%
    \normalfont B\kern-0.5em{\scshape i\kern-0.25em b}\kern-0.8em\TeX}}}

\makeatletter
\def\section{\@startsection{section}{1}{\z@}%
  {4.0ex plus 1.5ex minus 0.4ex}
  {1.4ex plus 0.3ex}
  {\centering\normalfont\scshape}}
\makeatother

\IEEEoverridecommandlockouts
\begin{document}

\title{Zamba2-VL Technical Report}

\newcommand{\corr}{\textsuperscript{*}}

\author{
\IEEEauthorblockN{Hassan Shapourian, 
Kasra Hejazi, 
Olabode M. Sule,
Beren Millidge}
\IEEEauthorblockA{Zyphra,\\ San Francisco, CA}
\IEEEauthorblockA{\texttt{\{hassan,kasra,bode,beren\}}\texttt{@zyphra.com}}
}

\maketitle

\setcounter{page}{1}

\begin{abstract}
We present Zamba2-VL, a suite of vision-language models built on Zamba2, a hybrid language-model architecture combining Mamba2 state-space layers with a small number of shared transformer blocks. Across a broad range of image understanding, reasoning, OCR, grounding, and counting benchmarks, Zamba2-VL is competitive with leading Transformer-based open-weight VLMs of comparable scale, including the Molmo2, Qwen3-VL, and InternVL3.5 families, and substantially outperforms prior SSM-based and hybrid VLMs such as VL-Mamba, Cobra, and mmMamba. Inheriting the near-linear prefill compute and small, near-constant recurrent state of its Zamba2 backbone, Zamba2-VL delivers roughly an order of magnitude lower time-to-first-token (TTFT) than these Transformer baselines at matched parameter scale, with the efficiency gap most pronounced at the smaller 1.2B and 2.7B scales most relevant to on-device and edge deployment. We release three models---1.2B, 2.7B, and 7B---together with inference code at \href{https://huggingface.co/collections/Zyphra/zamba2-vl}{https://huggingface.co/collections/Zyphra/zamba2-vl}.
\end{abstract}

\section{Introduction}

Vision-language models (VLMs) have become the dominant interface through which large pretrained models perceive and reason about the visual world, powering systems that range from multimodal chatbots and document understanding pipelines \cite{openai2023gpt4v, wei2025deepseekocrcontextsopticalcompression, poznanski2025olmocr2unittest} to medical assistants \cite{lasateam2025lingshugeneralistfoundationmodel, Chen-MedComplxReasoning}, computer-use agents \cite{wang2025uitars2technicalreportadvancing, wang2025opencua, gupta2026molmowebopenvisualweb}, embodied robotics \cite{pmlr-v270-kim25c, geminiroboticsteam2025geminiroboticsbringingai, ram2025from}, and autonomous driving \cite{sima2024drivelm, Shao2023LMDriveCE}. The dominant recipe is by now well established: a pretrained vision encoder \cite{Radford2021LearningTV, tschannen2025siglip, Kirillov_2023_ICCV, simeoni2025dinov3} feeds visual features through a lightweight connector into a pretrained large language model, which is then fine-tuned end-to-end on multimodal data. The connector itself has evolved from cross-attention modules \cite{alayrac2022flamingo} and learnable query bottlenecks \cite{BLIP-2-Li} to the now-standard vision connector module introduced by LLaVA \cite{liu2023visual} and adopted by Qwen3-VL \cite{bai2025qwen3vltechnicalreport}, InternVL3 \cite{zhu2025internvl3exploringadvancedtraining}, GLM4.5 \cite{vteam2026glm45vglm41vthinkingversatilemultimodal}, and Molmo \cite{deitke2025molmo}, among others. Despite differences in detail, almost every competitive open-weight or proprietary VLM \cite{openai2024gpt4technicalreport, anthropic2024claude35} shares this template, and essentially every one of them inherits a Transformer LLM at its core.

The choice of backbone has important consequences that are easy to overlook when talking about benchmark scores. Self-attention has quadratic compute in sequence length during prefill, and autoregressive decoding maintains a KV cache that grows linearly with every generated token.. While this is increasingly costly in the text-only regime, it becomes a hard constraint very rapidly for multimodal models. A single high-resolution image processed by modern tiling schemes such as AnyRes \cite{liu2024llavanext} or native-resolution variants based on 2D RoPE \cite{kexuefm-8397, kexuefm-10040, heo2024ropevit, wang2024qwen2, liu2026spiralrope} can contribute several thousand vision tokens; a short video clip routinely produces tens to hundreds of thousands. A range of mitigations has been proposed at the input side: visual token compression \cite{shang2024LLaVA-PruMerge, yang2025pvc, bolya2023tome} prunes, merges, or adaptively selects tokens before they reach the LLM, and `native' VLMs \cite{diao2026from} fold the vision encoder into the language model to allow joint end-to-end token reduction. These approaches reduce the symptom by feeding fewer tokens to attention, but they do not change the underlying scaling behavior of the backbone.

A more direct alternative is to change the backbone itself. State-space models, especially the selective SSM Mamba \cite{gu2023mamba} and its successor Mamba2 \cite{dao2024mamba2}, offer linear-time sequence processing and a constant-size recurrent state during autoregressive generation, replacing the growing Transformer KV cache with a fixed-dimensional state. Mamba2 formalizes these models through structured state space duality, relating selective SSMs to attention-like operators via structured semiseparable matrices and enabling more hardware-efficient implementations. On reported language-modeling benchmarks, these models are competitive with Transformers at small to medium scale and can provide substantially higher throughput in long-context generation. These efficiency properties have motivated SSM-based architectures throughout the vision stack. Vision backbones such as Vision Mamba (Vim) \cite{zhu2024vim} and VMamba \cite{liu2024vmamba} adapt selective scans to images using bidirectional or multi-directional scan patterns, addressing the mismatch between one-dimensional recurrence and two-dimensional visual structure, and report competitive or superior results against ViT/Swin-style baselines on classification, detection, and segmentation, particularly in high-resolution settings. On the VLM side, VL-Mamba \cite{qiao2024vlmamba} replaces the Transformer language backbone used in LLaVA-style models with a pretrained Mamba language model and introduces a multimodal connector with vision selective scan; Cobra \cite{zhao2024cobra} combines a Mamba language backbone with DINOv2 and SigLIP visual features and reports roughly $3$--$4\times$ faster inference than comparable efficient Transformer VLM baselines; Mamba-VL \cite{abouelenin2024mambavl} provides controlled comparisons against Transformer VLMs trained under matched conditions; and mmMamba \cite{liao2025mmmamba} distills decoder-only Transformer VLMs into pure-Mamba and hybrid Mamba--Transformer variants, achieving up to $20\times$ inference speedups at long contexts.

These efforts establish that SSMs are a viable substrate for vision-language modeling, but they also surface two patterns. The first is that pure-SSM VLMs based on early Mamba LLMs tend to lag attention-based counterparts on tasks that lean heavily on precise in-context lookup --- visual grounding, referring expressions, fine-grained retrieval --- where attention's content-addressable mechanism remains a relative strength \cite{abouelenin2024mambavl}. The second is that the strongest SSM-based VLMs to date have either been distilled from existing Transformer VLMs \cite{liao2025mmmamba} or built on relatively small open-source SSM LLMs, leaving open the question of how far a from-scratch SSM-family VLM can be pushed.

The Zamba2 series \cite{glorioso2024zamba2, glorioso2024zamba} was built to answer the LLM half of that question. Zamba2 interleaves a Mamba2 backbone with a small number of shared transformer blocks, each lightly specialized at its insertion point through LoRA projectors. This hybrid design is a deliberate response to the same trade-off the SSM-VLM literature has observed: the Mamba layers carry the bulk of the computation in linear time and constant state, while the shared attention layers provide precisely the in-context retrieval capability that pure-SSM models give up. The resulting 1.2B, 2.7B, and 7B models are competitive with the leading open-weight Transformer LLMs of their time at their respective scales while being substantially faster at generation and considerably cheaper to serve. Each of these properties --- linear-time prefill over long visual contexts, a fixed-size state during decoding, and preserved retrieval capability --- is exactly what a VLM workload demands.

In this report we present \textbf{Zamba2-VL}, a family of vision-language models built on the 1.2B, 2.7B, and 7B Zamba2 backbones. To our knowledge, Zamba2-VL is the strongest open hybrid SSM--Transformer VLM family released to date, and the first to demonstrate that the inference-efficiency advantages of hybrid state-space LLMs carry cleanly into the multimodal setting. Across image understanding, visual reasoning, OCR, grounding, and counting benchmarks, Zamba2-VL is competitive with leading Transformer-based VLMs of comparable scale --- including Molmo2 \cite{clark2026molmo2}, Qwen3-VL \cite{bai2025qwen3vltechnicalreport}, and InternVL3.5~\cite{zhu2025internvl3exploringadvancedtraining} --- while delivering order-of-magnitude inference speedups in the regimes where multimodal context is large.

Adapting the Zamba2 backbone to a vision-language setting requires choices that are not pinned down by the pretrained LLM alone, in particular the selection of the vision encoder and the composition of the training data mixture. We ablate these choices and report the configurations that proved most effective.

Beyond architectural advances, recent VLM work has shown that data quality and curation are at least as decisive as architectural choices for downstream performance \cite{deitke2025molmo, cho2025perceptionlm, an2025llavaonevision15, li2024llava}. Yet many otherwise-open VLMs continue to depend on synthetic data distilled from proprietary closed-weight models \cite{deitke2025molmo, clark2026molmo2}, which limits reproducibility and obscures the contribution of individual data sources. We treat data composition as a central design axis for Zamba2-VL: pre-training captions, instruction-tuning mixtures, and task-specific annotations are constructed and ablated alongside the architectural choices. Several remaining VLM challenges --- hallucination \cite{li-etal-2023-evaluating, kanade-ganu-2026-see, augustin2025dash}, edge deployment \cite{li2025eureka, VLM4Edge-Sharshar}, robustness \cite{zhang2025anyattack, wang2025advedm}, and 3D and multi-view understanding \cite{chen2025scene, SpatialVLM-Chen, HOU2026104314} --- fall outside the scope of this work, but the efficiency profile of Zamba2-VL is directly relevant to several of them, particularly deployment in latency- or memory-constrained settings.

Our main contributions are summarized as follows.

\begin{enumerate}
\item We present Zamba2-VL, the first family of open vision-language models built on a strong hybrid SSM--Transformer LLM backbone, and show that hybrid state-space LLMs are not merely viable but competitive substrates for VLMs at scales from 1.2B to 7B parameters.
\item We describe the architectural choices and training pipeline --- connector design, vision-token integration, data composition, and multi-stage curriculum --- required to extend a hybrid Mamba2--Transformer backbone into a fully capable VLM, and we ablate the design decisions that proved consequential.
\item We characterize Zamba2-VL's performance--efficiency profile: competitive accuracy against leading Transformer-based VLMs of comparable size on image understanding, reasoning, OCR, grounding, and counting benchmarks, combined with order-of-magnitude inference speedups inherited from the Zamba2 backbone. The efficiency advantage is most pronounced at the 1.2B and 2.7B scales, where the gap with Transformer VLMs in latency and serving cost is the largest.
\item We release Zamba2-VL at all three scales (1.2B, 2.7B, 7B), along with inference code, as a resource for the research and application community. Checkpoints are available at \href{https://huggingface.co/collections/Zyphra/zamba2-vl}{https://huggingface.co/collections/Zyphra/zamba2-vl} and inference code at \href{https://github.com/Zyphra/transformers/tree/zamba2-vl}{https://github.com/Zyphra/transformers/tree/zamba2-vl}.
\end{enumerate}

The remainder of this report is organized as follows. Section~\ref{sec:related-work} discusses related work on VLM design and on prior SSM-based vision and vision-language models in more detail. Section~\ref{sec:architecture} describes the Zamba2-VL architecture, including the vision encoder, connector, and integration with the Zamba2 hybrid backbone. Section~\ref{sec:training} details the data curation, alignment, and supervised fine-tuning pipeline. Section~\ref{sec:evaluation} reports benchmark results alongside inference-time efficiency measurements against comparable Transformer-based VLMs. Section~\ref{sec:ablations} presents ablation studies isolating the impact of key design choices. Section~\ref{sec:conclusions} concludes and outlines directions for future work.

\begin{figure*}[htbp]
    \centering
    \includegraphics[width=0.7\linewidth]{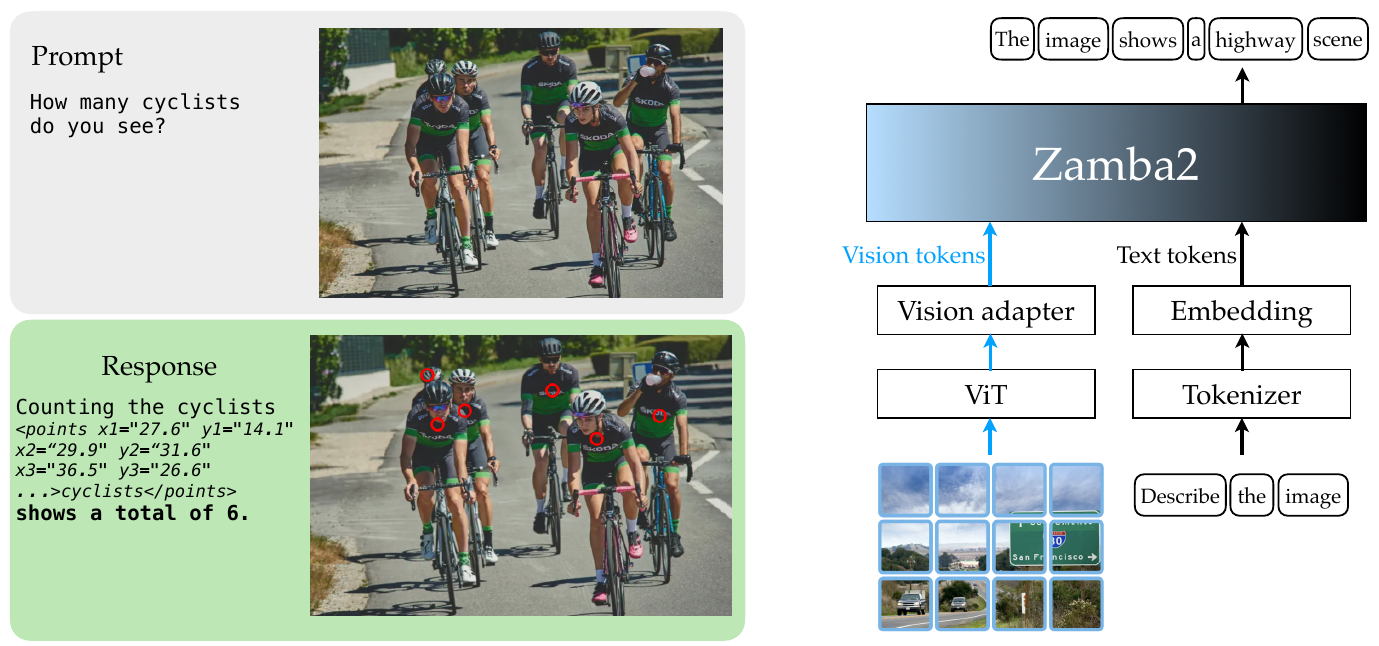}
    \caption{Left: A sample model response involving counting the objects by pointing to them. Right: Architecture of Zamba2-VL. The model uses Zamba2 as the LLM backbone and the Qwen2.5-VL vision encoder as the vision encoder, connected by a two-layer MLP adapter that projects image features into the language embedding space.}
    \label{fig:model-arch}
\end{figure*}

\section{Related Work}
\label{sec:related-work}

The space of open small-to-mid-scale VLMs has become increasingly crowded as the practical utility of multimodal models has become widely recognized. Strong open-weight families at the 1B--8B scale --- including LLaVA \cite{liu2023visual}, the Qwen-VL series \cite{bai2025qwen2.5, bai2025qwen3vltechnicalreport}, InternVL3 \cite{zhu2025internvl3exploringadvancedtraining}, GLM4.5V \cite{vteam2026glm45vglm41vthinkingversatilemultimodal}, Molmo \cite{deitke2025molmo, clark2026molmo2}, and PerceptionLM \cite{cho2025perceptionlm} --- have served as our quality benchmarks throughout the development of Zamba2-VL. Architecturally, these models share a common template: a pretrained vision encoder feeds visual features through a vision connector into a pretrained LLM, with the LLM fine-tuned to interpret the projected tokens. Variations among them lie mostly in vision-token handling rather than connector design --- AnyRes tiling for high-resolution inputs \cite{liu2024llavanext}, 2D RoPE for native-resolution processing \cite{kexuefm-8397, kexuefm-10040, heo2024ropevit, wang2024qwen2, liu2026spiralrope}, and visual token compression \cite{shang2024LLaVA-PruMerge, yang2025pvc, bolya2023tome}. We adopt the LLaVA-style MLP connector in Zamba2-VL.

Training pipelines across these models typically follow a multi-stage curriculum: connector alignment with the vision encoder and LLM frozen, supervised instruction tuning with all components updated jointly, and sometimes an RL post-training stage. Specific recipes differ in their stage composition and data mixtures --- LLaVA-OneVision \cite{li2024llava} inserts a high-quality knowledge stage, Molmo2 \cite{clark2026molmo2} adds long-context supervised fine tuning (SFT), PerceptionLM \cite{cho2025perceptionlm} relies on synthetic-data midtraining, and the Qwen-VL series \cite{bai2025qwen2.5, bai2025qwen3} trains its vision encoder from scratch --- but the common thread is a gradual increase in data complexity and quality across stages. Zamba2-VL adopts this conventional pattern and leaves RL post-training to future work.

A separate line of work scales state-space and hybrid SSM--Transformer models into the multimodal setting. The motivation mirrors that of the language-only case: SSMs such as Mamba \cite{gu2023mamba} and Mamba2 \cite{dao2024mamba2} offer linear-time inference and a fixed-size recurrent state in place of the growing KV cache of attention --- properties that become especially valuable when multimodal contexts contain thousands of vision tokens per image. On the vision-encoder side, Vision Mamba (Vim) \cite{zhu2024vim} and VMamba \cite{liu2024vmamba} apply bidirectional and multi-directional SSM scans to image patches, with MambaVision \cite{hatamizadeh2024mambavision} extending these ideas to hybrid attention--SSM image backbones. On the VLM side, VL-Mamba \cite{qiao2024vlmamba} swaps the LLaVA Transformer LLM for a pretrained Mamba and introduces a vision selective-scan connector; Cobra \cite{zhao2024cobra} couples Mamba with DINOv2 and SigLIP and reports $3$--$4\times$ inference throughput gains over Transformer VLMs of comparable size; Mamba-VL \cite{abouelenin2024mambavl} provides controlled head-to-head comparisons against Pythia-based Transformer VLMs trained on identical data; and mmMamba \cite{liao2025mmmamba} distills decoder-only Transformer VLMs into pure-Mamba and hybrid Mamba--Transformer variants, reporting up to $20\times$ speedups at long contexts. A consistent finding across these works, mirroring observations in the language-only regime \cite{jelassi2024repeat}, is that pure-SSM VLMs lag attention-based models on tasks requiring precise in-context lookup. Hybrid architectures address this directly: Jamba \cite{lieber2024jamba} interleaves Mamba, attention, and mixture-of-experts in a single LLM, and the Zamba and Zamba2 series \cite{glorioso2024zamba, glorioso2024zamba2} interleave a Mamba/Mamba2 backbone with shared transformer blocks augmented by per-insertion LoRA projectors. Zamba2-VL is, to our knowledge, the first VLM family built on a hybrid SSM--Transformer LLM of this kind.

Finally, VLM progress has been driven as much by data and evaluation infrastructure as by architectural innovation. Open training resources span image-text pairs for contrastive and caption-based pre-training \cite{Schuhmann-LAION-5B, Chen2015MicrosoftCC, sharma-etal-2018-conceptual, ShareGPT4V-Chen, zhu2024minigpt}, counting and VQA \cite{acharya2019tallyqa}, OCR \cite{SynthText-Gupta, ocr-vqa-mishra}, chart and figure understanding \cite{methani2020plotqa, kahou2017figureqa, kafle2018dvqa, yang2025effective}, object detection and visual grounding \cite{shao2019objects365, kuznetsova2020openimages, refCOCO-Mao}, and GUI and computer-use trajectories \cite{liu2024multiui, wu2024osatlas, GUIWorld-Lei, chen-etal-2025-guicourse}. The evaluation landscape has grown in parallel, with benchmarks covering general VQA and multi-discipline reasoning \cite{yue2024mmmu, yue-etal-2025-mmmu-pro}, STEM reasoning \cite{lu2023mathvista}, OCR \cite{liu2024ocrbench, poznanski2025olmocr2unittest}, chart understanding \cite{methani2020plotqa}, code generation from screenshots \cite{si-etal-2025-design2code}, video understanding \cite{hu2025videommmuevaluatingknowledgeacquisition, MVBench-Li}, GUI grounding and computer-use agents \cite{Screenspot-Pro-Li, xie2024osworld, he-etal-2024-webvoyager, rawles2025androidworld}, and hallucination and robustness \cite{li-etal-2023-evaluating, Hallusionbench-Guan, RBench-Zhao}. We report Zamba2-VL results on a representative subset of these in Section~\ref{sec:evaluation}.

\section{Model Architecture}
\label{sec:architecture}

Zamba2-VL adopts the LLaVA-style VLM architecture~\cite{liu2023visual} illustrated in Fig.~\ref{fig:model-arch}: an image preprocessor, a vision encoder, an MLP adapter that projects visual features into the LLM embedding space, and a decoder-only LLM that consumes the resulting interleaved sequence of vision and text tokens. We instantiate this template at three scales --- 1.2B, 2.7B, and 7B parameters --- using the corresponding Zamba2 LLMs~\cite{glorioso2024zamba2} as language backbones. Zamba2 is a hybrid architecture in which a Mamba2~\cite{dao2024mamba2} state-space backbone is interleaved with a small number of shared transformer blocks augmented by LoRA projectors~\cite{hu2021lora} which are not shared; the shared blocks supply the attention-style in-context retrieval capability that pure SSMs lack~\cite{jelassi2024repeat}, while the Mamba2 layers carry the bulk of the computation in linear time and with a fixed-size recurrent state. We refer the reader to~\cite{glorioso2024zamba2} for full backbone details.

For visual encoding we adopt the Vision Transformer from Qwen2.5-VL~\cite{bai2025qwen2.5}, motivated by its strong empirical performance in our setting. We attribute this in large part to its use of 2D Rotary Position Embeddings~\cite{kexuefm-8397, kexuefm-10040} applied directly to image patches and to its native dynamic-resolution processing~\cite{dehghani2023patch, wang2024qwen2}, which together avoid fixed-resolution distortions and preserve fine-grained spatial structure. For the LLM backbone we keep the position-embedding scheme of each Zamba2 variant unchanged --- standard 1D RoPE~\cite{su2024roformer} in the shared attention blocks of the 1.2B and 7B, and no rotary embedding in the 2.7B model (the latter owing to implementation and timing details during the original Zamba2 architecture search) --- rather than adopting a multimodal RoPE variant, since prior work suggests that such modifications require substantially more compute and data to yield consistent gains than was available in our training budget.

Each input image is resized so that its height and width are multiples of 28, with the aspect ratio preserved as closely as possible, and then fed to the ViT with a patch size of $14 \times 14$. A two-layer MLP adapter pools each $2 \times 2$ window of patch embeddings into a single vector and projects it into the LLM embedding space, simultaneously reducing the number of vision tokens by a factor of four and aligning their dimensionality with the text embedding space. The resulting vision tokens are interleaved with text tokens and consumed by the Zamba2 backbone in the usual autoregressive manner.

\section{Training}
\label{sec:training}

\begin{table}
\small
\setlength{\tabcolsep}{4pt}
\begin{minipage}{\linewidth}
    \centering
    \begin{tabular}{lcccc}
    \toprule
    \textbf{Stage} & \textbf{Training} & \textbf{Tokens} & \textbf{Max Len.} & \textbf{Max Res.} \\
    \midrule
    Alignment & Adapter & 230M & 800 & 0.3MP\\
    Pretraining & Full & 30B & 4k & 2.7MP \\
    Instruction Tuning & Full & 20B & 4k & 2.7MP \\
    \bottomrule
    \end{tabular}
    \caption{Training stages of Zamba2-VL. In Stage~1 (Alignment) we train only the MLP adapter on low-resolution image captioning data, with the loss computed over all text tokens. In Stage~2 (Pretraining) we unfreeze the full model and train on a heterogeneous multimodal mixture. Stage~3 (Instruction Tuning) is full-model training on curated instruction-following data. Across all stages, images are preserved at their native resolution up to a stage-specific cap, beyond which they are resized. From Stage~2 onward the loss is computed exclusively over answer tokens, so the model is supervised only on its own responses rather than on input context or questions.}
    \label{tab:train-stages}
\end{minipage}
\end{table}

Zamba2-VL is trained in three stages, summarized in Table~\ref{tab:train-stages}: adapter alignment, large-scale multimodal pretraining, and supervised instruction tuning. The training data mixture progressively shifts from short image captions in alignment, to a broad multimodal corpus in pretraining, to a curated instruction-following mixture in SFT. We describe each stage in turn, followed by the conventions that are shared across stages.

\emph{Chat template.} For pretraining and instruction tuning we use a lightweight chat template that brackets each image with dedicated vision-boundary tokens and delimits each conversational turn with the LLM tokenizer's existing beginning- and end-of-sequence markers \texttt{<s>} and \texttt{</s>}:
\begin{verbatim}
<|vision_start|><image><|vision_end|>
<s>user
question.</s>
<s>assistant
answer</s>
\end{verbatim}
\texttt{<|vision\_start|>} and \texttt{<|vision\_end|>} are newly introduced tokens that delineate image boundaries and allow the model to distinguish between the input image and the text span. Beyond these two, no additional role-delimiter tokens are introduced; turn boundaries are signaled by the tokenizer's native BOS/EOS, which keeps changes to the embedding table minimal.

\subsection{Stage 1: Alignment}
This stage initializes the vision-language interface in isolation. Only the MLP adapter is trained; both the vision encoder and the Zamba2 language backbone remain frozen. We use low-resolution image-caption pairs from LLaVA-ReCap-558K~\cite{llava_recap_558k}, cap input sequences at 800 tokens, and limit images to 0.3MP. The loss is computed over text tokens. The goal here is purely to obtain a good adapter initialization before the rest of the model is unlocked.

\subsection{Stage 2: Pretraining}
The full model is then trained jointly on a heterogeneous multimodal mixture totaling roughly 30B tokens. The mixture spans image captioning, general VQA, OCR and document understanding, chart and figure understanding, and visual grounding. Within this mixture we deliberately upsample document-understanding and OCR data relative to their natural proportions. This decision was driven by a controlled comparison: in LLaVA-NeXT--style ablation runs against a Llama-3.x reference model of comparable scale, the Zamba2 backbone matched the Transformer reference on general image-understanding benchmarks but lagged on text-heavy, document, and OCR tasks. The 4k cap is inherited from the context length at which the Zamba2 base models were pretrained~\cite{glorioso2024zamba2}; extending it would require a context-length extension stage on the LLM backbone first, which we leave to future work.

\subsection{Stage 3: Supervised Fine-Tuning}
The final stage performs instruction tuning on roughly 20B tokens of curated multimodal data, reusing the chat template and loss-masking scheme from pretraining. Compared to pretraining, the SFT mixture places greater weight on multi-turn conversation, instruction following, and grounding. Grounding is supervised in two formats. Pointing follows the XML format introduced in PixMo~\cite{deitke2025molmo}, in which each annotated point is emitted as an inline tag of the form \texttt{<point x="\_" y="\_" alt="label">label</point>} for single-point answers, and \texttt{<points x1="\_" y1="\_" x2="\_" y2="\_" ... alt="label">label</points>} when multiple points share the same label. The coordinates $x_i, y_i$ are expressed as relative percentages of image width and height, respectively, with one decimal place of precision and within the range $[0,100]$. Bounding-box supervision uses the plain coordinate format \texttt{[x1, y1, x2, y2]}, where coordinates are normalized to the range $[0,1]$ with two decimal places of precision, with no additional special tokens. In both formats the coordinates are resolution-independent, which keeps the targets consistent with the dynamic-resolution input pipeline regardless of the size at which an image is processed.

\emph{Loss masking.} From Stage 2 onward we compute cross-entropy only over answer tokens; images and questions act as conditioning context. This implicitly weights longer responses more heavily, but we found it preferable in practice to alternatives that include question or context tokens in the loss~\cite{shapourian2026zaya1}.

\emph{Effective batch size.} Because only answer tokens contribute to the loss, the effective gradient signal per optimizer step is substantially smaller than in equivalent text-only LLM training at the same nominal batch size. We compensate with a larger nominal batch size than is typical for LLM pretraining. The cross-entropy loss is computed per token and summed within each minibatch, with the per-step loss normalized by the number of answer tokens in that minibatch.

\begin{figure}
    \centering
    \includegraphics[width=\linewidth]{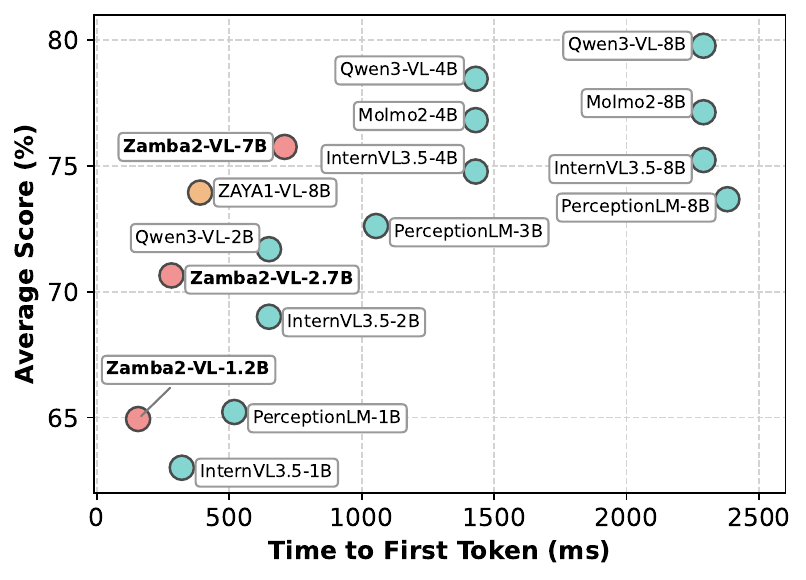}
    \caption{Average benchmark score vs.\ time to first token (TTFT) for Zamba2-VL and Transformer-based open-weight VLMs of comparable scale. TTFT is measured on a 32k-token prefill of the LLM backbone, where most inference compute is spent. Scores are averaged across the benchmarks of Table~\ref{tab:comprehensive-eval}. At every scale, Zamba2-VL achieves competitive accuracy at roughly an order of magnitude lower TTFT than the closest Transformer baseline.}
    \label{fig:scores-vs-ttft}
\end{figure}

\newcolumntype{Y}{>{\centering\arraybackslash}X}   
\newcolumntype{G}{@{\hspace{8pt}}}                 

\begin{table*}[!htbp]
\centering
\caption{Performance of Zamba2-VL on general vision evaluations. For DocVQA and InfoVQA we report scores from the original papers since the evaluation requires submission to conference website.}
\label{tab:comprehensive-eval}
\small
\setlength{\tabcolsep}{2pt}
\renewcommand{\arraystretch}{1.12}
\begin{minipage}{\linewidth}\centering

\begin{tabularx}{\linewidth}{l*{10}{Y}G Y Y G Y Y}
\toprule
& \multicolumn{6}{c}{{Chart, Diagram, and Document Understanding}}
  & \multicolumn{6}{c}{{Perception and Reasoning}}
  & \multicolumn{2}{c}{{Counting}}
  \\
\cmidrule(lr){2-7}\cmidrule(lr){8-13}\cmidrule(lr){14-15}
\textbf{Model} 
& \rotatebox{90}{\footnotesize\makecell[l]{\textbf{AI2D}\\(test)~\cite{kembhavi2016diagram}}}
& \rotatebox{90}{\footnotesize\makecell[l]{\textbf{ChartQA}\\(test)~\cite{masry2022chartqa}}}
& \rotatebox{90}{\footnotesize\makecell[l]{\textbf{DocVQA}\\(test)~\cite{mathew2021docvqa}}}
& \rotatebox{90}{\footnotesize\makecell[l]{\textbf{InfoVQA}\\(test)~\cite{mathew2022infographicvqa}}}
& \rotatebox{90}{\footnotesize\makecell[l]{\textbf{TextVQA}\\(val)~\cite{singh2019towards}}}
& \rotatebox{90}{\footnotesize\makecell[l]{\textbf{OCRBench}\\ \cite{liu2024ocrbench}}}
& \rotatebox{90}{\footnotesize\makecell[l]{\textbf{VQA v2.0}\\(val)~\cite{goyal2017making}}}
& \rotatebox{90}{\footnotesize\makecell[l]{\textbf{MathVista}\\(mini)~\cite{lu2023mathvista}}}
& \rotatebox{90}{\footnotesize\makecell[l]{\textbf{MMMU}\\(val)~\cite{yue2024mmmu}}}
& \rotatebox{90}{\footnotesize\makecell[l]{\textbf{SEED}\\(image)~\cite{li2023seed}}}
& \rotatebox{90}{\footnotesize\makecell[l]{\textbf{Blink}\\(val)~\cite{fu2024blink}}}
& \rotatebox{90}{\footnotesize\makecell[l]{\textbf{RealWorldQA}\\ \cite{realworldqa2024}}}
& \rotatebox{90}{\footnotesize\makecell[l]{\textbf{CountBenchQA}\\ \cite{beyer2024paligemma}}}
& \rotatebox{90}{\footnotesize\makecell[l]{\textbf{PixMoCount}\\(test)~\cite{deitke2025molmo}}}
 \\
\midrule

PerceptionLM-1B & 85.7 & 79.2 & 90.7 & 63.0 & 78.2 & 79.0 & 80.0 & 51.9 & 35.0 & 76.3 & 45.6 & 68.6 & 62.2 & 17.7 \\
InternVL3.5-1B & 81.2 & 78.0 & 85.6 & 60.5 & 71.1 & 79.2 & 69.6 & 52.9 & 40.1 & 72.5 & 43.4 & 56.9 & 58.3 & 32.8 \\
\textbf{Zamba2-VL-1.2B} & 81.5 & 77.6 & 87.4 & 60.7 & 71.9 & 71.4 & 78.0 & 48.7 & 32.4 & 71.1 & 43.2 & 65.9 & 56.9 & 62.5 \\

\midrule
InternVL3.5-2B & 88.6 & 81.6 & 89.4 & 70.8 & 76.5 & 83.4 & 73.6 & 61.4 & 49.9 & 75.2 & 51.3 & 61.6 & 70.0 & 32.8 \\
Qwen3-VL-2B & 86.2 & 78.7 & 93.3 & 72.4 & 79.9 & 84.1 & 78.8 & 51.8 & 40.9 & 74.8 & 53.2 & 66.0 & 87.9 & 55.7 \\
PerceptionLM-3B & 92.2 & 85.1 & 93.8 & 74.6 & 80.0 & 80.1 & 76.9 & 61.6 & 41.4 & 78.3 & 49.8 & 73.1 & 88.1 & 41.6 \\
Molmo2-4B & 93.8 & 86.1 & 87.8 & 78.6 & 83.1 & 62.0 & 85.3 & 56.5 & 48.8 & 78.0 & 63.5 & 73.8 & 91.2 & 87.0 \\
Qwen3-VL-4B & 91.8 & 81.8 & 95.3 & 80.3 & 81.5 & 84.1 & 80.7 & 63.6 & 51.4 & 77.3 & 63.2 & 71.0 & 87.3 & 89.2 \\
InternVL3.5-4B & 92.0 & 86.4 & 92.4 & 78.0 & 77.6 & 82.0 & 76.4 & 72.8 & 57.2 & 76.3 & 58.2 & 67.8 & 82.5 & 47.3 \\
\textbf{Zamba2-VL-2.7B} & 85.8 & 79.6 & 90.9 & 66.5 & 77.4 & 73.6 & 79.6 & 51.0 & 37.7 & 73.0 & 42.3 & 61.7 & 87.5 & 82.5 \\
\midrule

PerceptionLM-8B & 91.8 & 86.0 & 94.6 & 80.9 & 80.4 & 84.2 & 84.0 & 62.2 & 43.8 & 78.6 & 27.0 & 77.1 & 90.8 & 50.1 \\
Qwen3-VL-8B & 92.3 & 82.8 & 96.1 & 83.1 & 83.2 & 87.2 & 82.5 & 66.1 & 54.3 & 77.5 & 65.9 & 71.9 & 90.6 & 83.4 \\
InternVL3.5-8B & 92.5 & 86.7 & 92.3 & 79.1 & 77.7 & 83.8 & 78.6 & 73.9 & 58.0 & 77.2 & 59.8 & 67.4 & 81.1 & 45.2 \\
Molmo2-8B & 94.1 & 86.0 & 93.2 & 80.1 & 83.9 & 61.4 & 86.1 & 61.4 & 47.7 & 77.3 & 59.6 & 71.8 & 91.2 & 86.1 \\
\textbf{Zamba2-VL-7B} & 90.6 & 85.3 & 92.9 & 74.8 & 81.0 & 81.6 & 82.8 & 61.2 & 43.8 & 74.9 & 49.3 & 66.5 & 90.6 & 85.3 \\

\bottomrule
\end{tabularx}

\end{minipage}
\end{table*}

Throughout training, we use the AdamW optimizer for all model parameters, consistent with the optimizer used during the base models' pretraining.

\subsection{Data}
\label{sec:data}

Our training corpus is assembled by curating and mixing a broad range of open-source datasets, drawing inspiration from the data strategies developed in PerceptionLM~\cite{cho2025perceptionlm}, Idefics3~\cite{laurencon2024building}, and Molmo~\cite{deitke2025molmo}. We organize the corpus into high-level capability categories --- general image understanding and captioning, document and OCR, grounding and perception, image-grounded question answering, multimodal reasoning, and text-only --- and vary their proportions across the three training stages. Pretraining is dominated by general image understanding, captioning, and document data, which together account for most of the token budget at this stage. SFT shifts emphasis toward higher-quality multimodal samples, with greater weight on grounding (especially bounding-box and pointing supervision) and multi-step multimodal reasoning. Given the aggregate size of the mixture, we stream all data online during training via Mosaic Streaming~\cite{mosaicml_streaming}.

For image understanding and captioning we draw most heavily on FineVision~\cite{wiedmann2025finevision}, a curated 24M-sample corpus assembled from over 200 sources with deduplication and decontamination against 66 benchmarks, as well as the highly detailed human-annotated captions of PixMo-Cap~\cite{deitke2025molmo},  the instruction-following data of MAmmoTH-VL~\cite{guo2024mammothvl}, and the interleaved data of M4-Instruct~\cite{li2024llavainterleavem4}. Question-answering capability is supported by The Cauldron~\cite{laurencon2024matters} alongside the standard set of academic VQA, OCR, and chart benchmarks --- VQAv2~\cite{goyal2017making}, OK-VQA~\cite{marino2019ok}, TextVQA~\cite{singh2019towards}, AI2D~\cite{kembhavi2016diagram}, ChartQA~\cite{masry2022chartqa}, DocVQA~\cite{mathew2021docvqa}, InfographicVQA~\cite{mathew2022infographicvqa}, A-OKVQA~\cite{schwenk2022okvqa}, ScienceQA~\cite{lu2022learn}, TabMWP~\cite{lu2022dynamic}, TallyQA~\cite{acharya2019tallyqa}, DVQA~\cite{kafle2018dvqa}, FigureQA~\cite{kahou2017figureqa}, and PlotQA~\cite{methani2020plotqa} --- and supplemented by ArxivQA~\cite{li2024multimodalarxiv} for scientific-figure reasoning, UReader~\cite{ye2023ureader} for unified text-rich understanding, and SynCLock~\cite{yang2022its} for analog-clock reading.

Document and OCR data is drawn from the PDFA dataset~\cite{montalvo2024pdfa}, the UCSF Industry Documents Library~\cite{montalvo2024ucsf}, DocMatix~\cite{laurencon2024building}, PixMo-Docs~\cite{deitke2025molmo}, UniChart~\cite{masry2023unichart}, and ECD-10K~\cite{yang2025effective}, with FineVision contributing additional document and chart subsets. Grounding and perception data combine 2D pointing and counting from PixMo-Point and PixMo-Count~\cite{deitke2025molmo}; UI and GUI grounding from MultiUI~\cite{liu2024multiui}, OS-Atlas~\cite{wu2024osatlas}, UGround~\cite{gou2024uground}, AutoGUI~\cite{li2025autogui}, and Aria-UI~\cite{yang2024ariaui}; and detection-style bounding-box supervision from OpenImages~\cite{kuznetsova2020openimages} and Objects365~\cite{shao2019objects365}. Multimodal reasoning is supported by chain-of-thought data in M$^3$CoT~\cite{chen2024m3cot}, geometric reasoning in Geometry3K~\cite{lu2021intergps} and Geo170K~\cite{gao2023gllava}, and broader STEM reasoning in ViRL39K~\cite{wang2025vl}. Finally, we included a small text-only fraction of GSM8K~\cite{cobbe2021gsm8k} for grade-school mathematical reasoning and to maintain text only performance.

\section{Evaluation}
\label{sec:evaluation}

We evaluate Zamba2-VL on a broad suite of vision-language benchmarks spanning chart, diagram, and document understanding (AI2D, ChartQA, DocVQA, InfoVQA, TextVQA, OCRBench), general perception and reasoning (VQAv2, MathVista, MMMU, SEED, Blink, RealWorldQA), and visual counting (CountBenchQA, PixMoCount). Results are summarized in Table~\ref{tab:comprehensive-eval}. This selection is designed to probe the major capability axes of modern VLMs: text recognition and visually situated reading, scene-level perception, multi-step multimodal reasoning, and precise object enumeration.

To make comparisons meaningful, we group the table into three parameter bands --- approximately 1B, 2--4B, and 7--8B --- and compare each Zamba2-VL model against the strongest contemporary open-weight Transformer-based VLMs in its band: PerceptionLM~\cite{cho2025perceptionlm}, InternVL3.5~\cite{zhu2025internvl3exploringadvancedtraining}, Qwen3-VL~\cite{bai2025qwen3vltechnicalreport}, and Molmo2~\cite{clark2026molmo2}. Every baseline in the table is built on a dense Transformer LLM backbone; Zamba2-VL is the only entry whose language model is hybrid SSM--Transformer. Direct comparisons against other SSM-based VLMs --- VL-Mamba, Cobra, mmMamba, and the Mamba-VL/Pythia-VL pairs --- are deferred to Appendix~\ref{app:ssm-comparison}, since the public evaluation data for those models is sparse and limited to a small set of older benchmarks (AI2D, TextVQA, VQAv2, POPE, GQA), several of which (notably POPE and GQA) are known to contain incorrect ground-truth annotations and are no longer considered reliable signals of VLM capability. Within that restricted comparison set, Zamba2-VL substantially outperforms prior SSM-based VLMs at every scale we report. We note that several of the 2--4B-class baselines (Molmo2-4B, Qwen3-VL-4B, InternVL3.5-4B) possess 40--50\% more parameters than Zamba2-VL-2.7B, which makes the comparison within that band more demanding for Zamba2-VL compared to the others.

The overall picture is that Zamba2-VL is broadly competitive with the best Transformer-based open-weight models in each band, and notably strong on counting and on document- and chart-style benchmarks --- consistent with the document- and OCR-upsampling strategy described in Section~\ref{sec:data}. The clearest standout is visual counting. On PixMoCount, Zamba2-VL-1.2B reaches 62.5, nearly double the score of the Transformer-based InternVL3.5-1B (32.8) and more than three times that of PerceptionLM-1B (17.7), despite a comparable or smaller parameter budget. The 2.7B and 7B models continue this trend, with PixMoCount scores of 82.5 and 85.3 respectively that place them at or near the top of their bands, and CountBenchQA scores of 87.5 and 90.6 that are competitive with directly comparable Transformer baselines. Document and chart understanding hold up similarly: Zamba2-VL is competitive on DocVQA, ChartQA, and TextVQA in every band, and the gap to the strongest Transformer baselines on these benchmarks is consistently smaller than the gap on reasoning-heavy benchmarks such as MMMU and MathVista, where Zamba2-VL trails. We view this profile --- strong perception and grounding, lagging multi-step reasoning --- as consistent with a training pipeline that emphasizes capability acquisition over RL-style reasoning specialization, which we leave to future work.

A second pattern is that Zamba2-VL's relative position improves with scale. The 1.2B model is competitive but trails the leading Transformer 1B baselines on roughly half of the benchmarks; the 2.7B model is broadly mid-pack within a band that includes Transformer models 40--50\% larger; and the 7B model is consistently competitive with the strongest 7--8B Transformer open-weight VLMs across nearly every category. This is the same scale-dependent narrowing we observed in the controlled ablation of Section~\ref{sec:ablations}, where the average gap between Zamba2 and a Llama-3.x Transformer reference shrank from $\sim$5 points at 1B to under one point at 7B. This trend is consistent with a growing body of evidence that hybrid architectures interleaving linear-time and full-attention layers can match or exceed pure-attention models of comparable scale: in the language-only setting, recent hybrids such as Kimi-Linear~\cite{team2025kimilinear} and Gemma-3~\cite{gemmateam2025gemma3} report parity with or improvements over full-attention baselines under matched training, and in the multimodal setting, hybrid Mamba--Transformer VLMs such as MaTVLM~\cite{li2025matvlm} and the long-video models Vamba~\cite{ren2025vamba} and TimeViper~\cite{wang2025timeviper} demonstrate that a small number of attention layers is sufficient to retain competitive accuracy while substantially reducing inference cost. Our results add to this picture from the perspective of a from-scratch open hybrid VLM family trained across three scales, and suggest that the hybrid SSM--Transformer recipe matches the capability of dense Transformers at scale while retaining its inference-efficiency advantages.

The benchmark scores alone tell only half the story. Figure~\ref{fig:scores-vs-ttft} plots the average benchmark score against time to first token (TTFT) on a 32k-token prefill --- the latency-critical phase of inference, where attention's quadratic cost is felt most acutely and where the predominantly linear-time prefill of the Zamba2 backbone yields its largest absolute advantage. At every parameter scale, Zamba2-VL occupies the upper-left region of the plot, achieving competitive accuracy at roughly an order of magnitude lower TTFT than the closest-scoring Transformer baseline. No Transformer-based VLM in the comparison set matches the score of any Zamba2-VL model at comparable latency, and the latency gap is at least an order of magnitude in the regimes that matter most for serving.

Taken together, these results show that the hybrid SSM--Transformer recipe is competitive with dense Transformer VLMs of comparable scale on standard vision-language benchmarks while delivering inference latency that is up to an order of magnitude lower. We see this as direct evidence that hybrid SSM--Transformer LLMs are a practical alternative to dense Transformers as VLM backbones, particularly when memory or latency is a binding constraint at deployment.

\emph{Reproducibility.}
All results except DocVQA and InfoVQA in Table~\ref{tab:comprehensive-eval} are reproduced through a single pipeline, VLMEvalKit~\cite{duan2024vlmevalkit}, in order to eliminate scoring inconsistencies that arise from differing prompting conventions across model releases. For DocVQA and InfoVQA, we report the scores published in each model's original paper, since the official test sets require submission to an external evaluation server. A small number of reproduced scores diverge mildly from those published in the original references; where this occurs we have verified that the divergence is not driven by a systematic prompting issue on our end. For PixMoCount, the public test set nominally contains 540 examples; of these, 531 were successfully retrieved by VLMEvalKit in our environment, and all reported PixMoCount numbers in the table are computed on this 531-image subset across every model evaluated, so the comparison remains internally consistent. Finally, we note one systematic evaluation-time asymmetry that affects InfoVQA in particular. Zamba2-VL is trained at 4k context length with a per-image budget of 3.4k vision tokens, and we retain this cap at inference time. InfoVQA contains many very high-resolution infographics whose native resolution would, for the Transformer baselines, expand to many more vision tokens; in our setup these large images are downscaled to fit the 3.4k-token budget, which likely reduces accuracy on text-dense regions. The other benchmarks in the suite are largely unaffected because their typical image resolution falls below this budget.

\section{Ablation Studies}
\label{sec:ablations}

We validate two of the key design choices behind Zamba2-VL through controlled ablations: the choice of vision encoder and the choice of language backbone. In both cases, ablations begin from the aligned checkpoint produced by Stage~1 (see Section ~\ref{sec:training}) for each vision encoder and run the pretraining stage on the LLaVA-NeXT 790K ablation dataset~\cite{LLaVA-NeXT-Data}. Training uses a batch size of 128, a maximum image budget of $3400$ vision tokens, and a maximum text length of 600 tokens. All other training details follow Stage~2 of the main pipeline.

We evaluate on six image-understanding and perception benchmarks --- AI2D, ChartQA, DocVQA, InfographicVQA, TextVQA, and RealWorldQA --- and summarize overall performance through the unweighted mean of these scores, which we report as ``Img. Und. Avg.''

\begin{table}[t]
\centering
\small
\setlength{\tabcolsep}{6pt}
\renewcommand{\arraystretch}{1.12}
\begin{tabular}{lc}
\toprule
\textbf{ViT}  & \textbf{Img. Und. Avg.} \\
\midrule
CLIP                  & 61.87 \\
SigLIP-2              & 66.18 \\
\rowcolor{gray!20}
Qwen2.5-VL ViT        & 73.14 \\
\bottomrule
\end{tabular}
\caption{Choice of vision encoder, holding the language backbone fixed at Zamba2-2.7B.}
\label{tab:vit-ablation}
\end{table}

\paragraph{Vision encoder}
Table~\ref{tab:vit-ablation} compares three vision encoders --- CLIP~\cite{Radford2021LearningTV}, SigLIP-2~\cite{tschannen2025siglip}, and the ViT from Qwen2.5-VL~\cite{bai2025qwen2.5} --- with the language backbone held fixed at Zamba2-2.7B. The Qwen2.5-VL ViT outperforms SigLIP-2 by a large margin. We attribute the gap primarily to two properties of the Qwen2.5-VL ViT discussed in Section~\ref{sec:architecture}: its use of 2D RoPE applied directly to image patches, and its native dynamic-resolution processing, both of which preserve fine-grained spatial structure on the text-heavy and document-style images that dominate the evaluation suite. We adopt the Qwen2.5-VL ViT as the vision encoder for all Zamba2-VL models.

\begin{table*}[t]
\centering
\small
\setlength{\tabcolsep}{6pt}
\renewcommand{\arraystretch}{1.12}
\begin{tabular}{lccccccc}
\toprule
\textbf{Backbone} & \textbf{Average}
& \footnotesize\makecell[c]{\textbf{AI2D}\\(test)}
& \footnotesize\makecell[c]{\textbf{ChartQA}\\(test)}
& \footnotesize\makecell[c]{\textbf{DocVQA}\\(val)}
& \footnotesize\makecell[c]{\textbf{InfoVQA}\\(val)}
& \footnotesize\makecell[c]{\textbf{TextVQA}\\(val)}
& \textbf{RealWorldQA} \\
\midrule
Llama-3.2-1B  & 61.92 & 64.5 & 65.2 & 76.0 & 38.0 & 65.9 & 48.5 \\
Zamba2-1.2B   & 56.44 & 63.9 & 62.6 & 67.4 & 25.9 & 62.4 & 50.1 \\
\midrule
Llama-3.2-3B  & 65.94 & 62.6 & 67.9 & 82.2 & 47.2 & 69.8 & 48.0 \\
Zamba2-2.7B   & 54.82 & 64.0 & 64.1 & 62.5 & 22.7 & 60.8 & 51.5 \\
\midrule
Llama-3.1-8B  & 73.08 & 83.5 & 74.8 & 85.4 & 48.9 & 72.8 & 60.0 \\
Zamba2-7B     & 72.64 & 83.6 & 74.1 & 85.0 & 48.3 & 72.2 & 60.6 \\
\bottomrule
\end{tabular}
\caption{Language backbone ablation: Zamba2 hybrid SSM--Transformer versus Llama-3.x Transformer reference at three roughly comparable scales, with the Qwen2.5-VL ViT held fixed across all rows.}
\label{tab:zamba-ablation}
\end{table*}

\paragraph{Language backbone: Mamba vs.\ Transformer}
Table~\ref{tab:zamba-ablation} compares Zamba2 hybrid backbones against Llama-3.2~\cite{meta2024llama32} (for the 1B and 3B references) and Llama-3.1~\cite{grattafiori2024llama3} (for the 8B reference) Transformer baselines of comparable scale, with the Qwen2.5-VL ViT held fixed. Three patterns emerge. First, the gap between the two architectures closes with scale: Llama-3.2-1B leads Zamba2-1.2B by 5.5 points on average, Llama-3.2-3B leads Zamba2-2.7B by 11 points, and at the largest scale Zamba2-7B is nearly indistinguishable from Llama-3.1-8B (72.64 vs.\ 73.08). At 7B, the hybrid SSM--Transformer backbone matches a strong Transformer of comparable size while retaining the linear-time inference and constant-state decoding properties of Mamba2.

Second, the per-benchmark breakdown reveals a clear capability profile rather than a uniform deficit. Zamba2 backbones are competitive or stronger on AI2D, ChartQA, TextVQA, and especially RealWorldQA, where Zamba2 outperforms the Llama reference at all three scales. The gap concentrates in DocVQA and InfographicVQA --- the two most text-dense, document-heavy benchmarks in the suite --- which we attribute to the relative scarcity of long, text-rich documents in the LLaVA-NeXT 790K ablation mixture. This observation directly motivates the document- and OCR-upsampling strategy described in Section~\ref{sec:data}, which is applied throughout the full Zamba2-VL pretraining mixture.

Third, the 2.7B variant is the weakest performer in absolute terms within the Zamba2 family on this benchmark suite, slightly underperforming the 1.2B model on average. We suspect this reflects a combination of factors specific to that variant --- notably the absence of rotary position embeddings in its shared attention blocks (Section~\ref{sec:architecture}) --- but emphasize that the effect is benchmark-suite-specific: the full Zamba2-VL evaluation results in Section~\ref{sec:evaluation} show that this gap largely closes once the full training pipeline, including the document- and OCR-upsampled mixture and the longer training horizon, is applied.

\section{Conclusions}
\label{sec:conclusions}

We presented Zamba2-VL, a family of vision-language models built on the Zamba2~\cite{glorioso2024zamba2} hybrid SSM--Transformer LLM at three scales (1.2B, 2.7B, and 7B). Across image understanding, OCR, grounding, and reasoning benchmarks, Zamba2-VL is competitive with leading open-weight Transformer-based VLMs of comparable size, while inheriting the inference-time efficiency advantages of the underlying Zamba2 backbones --- near-linear prefill compute over long visual contexts, a small, near-constant recurrent state in place of a growing KV cache during decoding, and the resulting headroom for deployment in latency- and memory-constrained settings. These gains were achieved with a total multimodal training budget of roughly 50B tokens, a fraction of the data used by the leading VLM families, suggesting that hybrid SSM--Transformer backbones can be competitive substrates for multimodal modeling without commensurate increases in data or compute. By open-sourcing Zamba2-VL at all three scales together with inference code, we hope to provide a practical foundation for researchers and practitioners to build upon.

Two empirical findings from the development of Zamba2-VL are worth highlighting. First, when compared to a Llama-3.x reference of comparable scale under matched training conditions, Zamba2 backbones match or exceed the Transformer baseline on most image-understanding categories but lag on text-dense document and OCR tasks; deliberately upsampling document and OCR data in the pretraining mixture closes this gap. Second, the inference efficiency advantages of Zamba2 are most pronounced precisely at the 1.2B and 2.7B scales, where latency and serving cost are the dominant deployment constraints --- making the smaller members of the Zamba2-VL family particularly well suited to on-device and edge applications.

Several directions follow naturally from this work. The most immediate direction concerns context length: we trained Zamba2-VL on the standard 4k-context Zamba2 base models, which caps the resolution and number of images that fit in a single context. Running the backbones through a context-length extension stage before continuing Zamba2-VL training is a natural next step, and is also the regime where the linear-time-inference advantages of the hybrid backbone matter most. Other promising directions include integrating a multimodal RoPE variant into the shared attention blocks of the Zamba2 backbone, scaling the training corpus, extending Zamba2-VL to video, and adding a reinforcement-learning post-training stage.

More broadly, Zamba2-VL is one demonstration of a more general claim: that the efficiency profile of hybrid sub-quadratic LLMs translates productively into the multimodal regime, and that the in-context retrieval limitations of purely recurrent models can be largely sidestepped through carefully placed shared attention blocks. The hybrid recipe is not unique to Mamba: recent linear-attention variants including Gated DeltaNet~\cite{yang2025gateddeltanet}, DeltaNet~\cite{yang2024deltanet}, and gated linear attention~\cite{yang2024gla} offer the same linear-time, constant-state inference profile and have themselves been combined with attention layers --- for instance, in Qwen3-Next's hybrid Gated-DeltaNet/attention design --- to similar effect. We view Zamba2-VL as evidence that the broader hybrid path, in which a sub-quadratic recurrence does the bulk of the work and a small amount of attention preserves recall, is a viable alternative to dense Transformer scaling for capable, deployable multimodal systems.

\section*{Acknowledgements}

We thank our colleagues at Zyphra; in particular, Quentin Anthony for running time-to-first-token experiments on LLM backbones, and Paolo Glorioso, Anthony Ndirango, Yury Tokpanov, and Robert Washbourne for insightful discussions. We thank Danny Martinelli and Paul White for their help with the Zamba2-VL public release.

\bibliographystyle{unsrt}
\bibliography{refs.bib}

\clearpage
\newpage

\appendices

\section{Comparison with other SSM-based VLMs}
\label{app:ssm-comparison}

\begin{table}[htb]
\centering
\small
\setlength{\tabcolsep}{3.5pt}
\begin{tabular}{lccccc}
\toprule
\textbf{Model}
& \footnotesize\makecell[c]{\textbf{AI2D}\\(test)}
& \footnotesize\makecell[c]{\textbf{TextVQA}\\(val)}
& \footnotesize\makecell[c]{\textbf{VQA v2.0}\\(val)}
& \footnotesize\makecell[c]{\textbf{POPE}\\(test)}
& \footnotesize\makecell[c]{\textbf{GQA}\\(test-dev)}
 \\
\midrule
Pythia-VL-1B           & 77.6 & 35.2 & 72.3 & \textbf{86.8} & 53.8 \\
Mamba-VL-0.8B          & 79.3 & 40.7 & 71.7 & \textbf{86.8} & 55.0 \\
Pythia-VL-1.4B         & 79.3 & 37.5 & 73.6 & 86.4 & 57.0 \\
Mamba-VL-1.4B          & 80.9 & 41.3 & 74.5 & 85.3 & \textbf{58.4} \\
\textbf{Zamba2-VL-1.2B}    & \textbf{81.5} & \textbf{71.9} & \textbf{78.0} & 84.2   & 57.9   \\
\midrule
Pythia-VL-2.8B         & 81.6 & 39.1 & 75.1 & 86.9 & 59.8 \\
Mamba-VL-2.8B          & 83.7 & 42.1 & 76.1 & 87.3 & 60.4 \\
VL-Mamba2.8B          & --   & --   & 76.6 & 84.4 & 56.2 \\
mmMamba2.7B           & --   & 55.1 & --   & 86.7 & 59.3 \\
Cobra-3.5B             & --   & 58.2 & 77.8 & \textbf{88.4} & \textbf{62.3} \\
\textbf{Zamba2-VL-2.7B}       & \textbf{85.8} & \textbf{77.4} & \textbf{79.6} & 87.2   & 58.3   \\
\midrule
Cobra-8B               & --   & 59.5 & 79.2 & 87.6 & \textbf{63.9} \\
\textbf{Zamba2-VL-7B}  & \textbf{90.6} & \textbf{81.0} & \textbf{82.8} & \textbf{88.6}   & 60.2   \\
\bottomrule
\end{tabular}
\caption{Comparison of Zamba2-VL against other SSM-based VLMs on a small set of benchmarks. Models are grouped by language-backbone scale into three bands: $\sim$1B (top), 2--4B (middle), and 7--8B (bottom). Bold indicates the highest score within each band on a given benchmark; ``--'' denotes results not reported by the source paper. Note that several of these benchmarks (notably POPE and GQA) contain known ground-truth issues and are not widely used as a signal of VLM capability in more recent papers; we include the table for completeness, since they are the benchmarks on which other SSM-based VLMs have publicly reported results.}
\label{tab:ssm-comparison}
\end{table}

\end{document}